\documentclass[11pt]{article}

\usepackage{acl}
\usepackage{times}
\usepackage{latexsym}
\usepackage[T1]{fontenc}
\usepackage[utf8]{inputenc}
\usepackage{microtype}
\usepackage{inconsolata}

\usepackage{needspace}
\usepackage{tabularx}
\usepackage{enumitem}
\usepackage{ragged2e}
\usepackage{booktabs}
\usepackage{array}
\usepackage{url}
\usepackage{graphicx}
\usepackage{amsmath,amssymb,amsfonts}
\usepackage{listings}
\usepackage{placeins}
\usepackage{xcolor}
\usepackage{mdframed}
\usepackage{xspace}

\definecolor{graybg}{HTML}{F5F5F5}

\newcommand\boxedname{Prompt\xspace}
\newcounter{prompt}
\newenvironment{prompt}[1][]{
    \refstepcounter{prompt}
    \begin{mdframed}[
        backgroundcolor=graybg,
        linecolor=black!50,
        linewidth=0.4pt,
        innerleftmargin=5pt,
        innerrightmargin=5pt,
        innertopmargin=4pt,
        innerbottommargin=4pt,
        skipabove=5pt,
        skipbelow=5pt,
        frametitle={\textbf{\boxedname\ \theprompt:} #1},
        frametitlefont=\footnotesize,
        frametitlerule=true
    ]
    \footnotesize\ttfamily
    \raggedright
    \setlength{\parindent}{0pt}
    \setlength{\parskip}{3pt}
}{
    \end{mdframed}
}

\lstdefinestyle{jsonstyle}{
    basicstyle=\ttfamily\footnotesize,
    backgroundcolor=\color{graybg},
    frame=single,
    rulecolor=\color{black!50},
    framerule=0.4pt,
    framesep=5pt,
    breaklines=true,
    breakatwhitespace=true,
    columns=fullflexible,
    showstringspaces=false,
    keepspaces=true,
    xleftmargin=0pt,
    xrightmargin=0pt,
    aboveskip=4pt,
    belowskip=4pt
}

\lstdefinestyle{reportstyle}{
    basicstyle=\ttfamily\fontsize{7}{8.2}\selectfont,
    breaklines=true,
    breakatwhitespace=true,
    columns=fullflexible,
    showstringspaces=false,
    keepspaces=true,
    frame=single,
    framerule=0.3pt,
    linewidth=\columnwidth,
    xleftmargin=0pt,
    xrightmargin=0pt,
    aboveskip=4pt,
    belowskip=4pt,
    tabsize=2
}

\title{
Trust but Verify:
Evidence-Linked Multi-Agent Clinical Information Extraction in Pathology
}

\author{
Yufan Wang\textsuperscript{1} \quad
Anit Kumar Sahu\textsuperscript{5} \quad
Yan Fei Ng\textsuperscript{2} \quad
Daniel Kang\textsuperscript{1}
\\[-0.1em]
Shayan Vassef\textsuperscript{1} \quad
Soorya Ram Shimgekar\textsuperscript{1} \quad
Koustuv Saha\textsuperscript{4}
\\[-0.1em]
Piyum Zonooz\textsuperscript{1} \quad
Navin Kumar\textsuperscript{1} \quad
Chee Leong Cheng\textsuperscript{2,3} \quad
Li Yan Khor\textsuperscript{2,3}
\\[0.2em]
\normalfont
\textsuperscript{1}Nimblemind.ai, USA
\\[-0.1em]
\textsuperscript{2}Department of Anatomical Pathology, Singapore General Hospital, Singapore
\\[-0.1em]
\textsuperscript{3}Duke--NUS Medical School, Singapore
\\[-0.1em]
\textsuperscript{4}University of Illinois Urbana-Champaign, USA
\\[-0.1em]
\textsuperscript{5}Independent Researcher
}

\begin{document}
\maketitle

\begin{abstract}
Clinical feature extraction from pathology reports is challenging because relevant evidence may be distributed across coded and narrative fields and depend on specimen attribution, negation, ancillary findings, and diagnostic context.
We retrospectively evaluated the NimbleMind Multi-Agent System (nMAS), a configurable workflow that separates clinician-defined field specifications from extraction models and returns report-level predictions with source-linked evidence. The study included 54 dummy gastric biopsy pathology reports from Singapore and four binary target fields, yielding 216 feature-case decisions.
nMAS correctly classified 213 of 216 decisions (98.61\%), and all evidence spans associated with correct predictions occurred verbatim in the corresponding source reports. All three errors occurred in the two context-dependent \textit{H. pylori}-related fields requiring negation handling or diagnostic attribution.
A single-model UMA-style comparator produced the similar label-level performance and error pattern. These findings do not demonstrate predictive superiority for the multi-agent architecture.Rather, the contribution of nMAS lies in workflow integration and traceability through configurable field specifications, complexity-based routing, report-level aggregation, and source-text validation within a clinician-reviewable workflow. Larger multi-institutional studies should assess generalizability, semantic evidence quality, adaptation effort, and clinician verification time.
\end{abstract}

\section{Introduction}

Clinical feature extraction converts clinically meaningful information in narrative documents into reusable structured variables, but it remains a persistent bottleneck across healthcare. Information needed for care, research, quality improvement, and cohort construction is often embedded in heterogeneous notes and reports, requiring clinicians or trained abstractors to review the same documents manually \cite{wang2018clinicalie,bian2026learningutility}. The burden can remain substantial even after automation: in one breast pathology validation study, 49.1\% of reports were still flagged for manual review and 30.8\% were incorrectly coded \cite{wieneke2015validation}. At an illustrative review time of five minutes per report, screening 1,000 reports would require 83.3 staff-hours, or about USD~6,250 at a rounded hourly cost of USD~75 \cite{jobstreet2026crc,moh2025healthcare_salaries,salaryexpert2026pathologist,wise2026sgdusd}.

Pathology extraction differs from generic clinical text classification in several concrete ways.
A single pathology report may describe multiple specimens, requiring each finding to be attributed to the correct specimen rather than detected anywhere in the document.
Relevant evidence may also be distributed across coded fields, specimen labels, final diagnoses, microscopic descriptions, and ancillary-test comments, requiring preservation of both section and specimen context.
In addition, the extractor must distinguish affirmative, negated, and uncertain mentions and determine whether findings are explicitly linked diagnostically.
Prior pathology NLP studies have documented substantial variation in report wording and reduced performance when relevant findings are rare, inconsistently expressed, or distributed across multiple report components \cite{buckley2012feasibility,wieneke2015validation}.

These challenges are concrete in gastric pathology.
About 31\% of the Singapore population was estimated to have evidence of \textit{Helicobacter pylori} infection, and persistent infection is associated with chronic active gastritis, peptic ulcer disease, and gastric cancer risk \cite{fock1997hpylori_singapore,malfertheiner2022maastricht,chey2024acg,lee2016eradication}.
Reliable identification of biopsy-confirmed cases can support treatment review, audit, research, and prevention programs \cite{chew2017hpylori_singapore,ang2021hpylori_strategy_singapore,ang2022clinical_audit_hpylori_sg}.
However, ``Helicobacter organisms are identified'' and ``No Helicobacter organisms are identified'' contain nearly identical terminology but require opposite labels.
Moreover, organism positivity alone does not establish \textit{H. pylori}-associated gastritis unless the report explicitly attributes the gastritis to the organism \cite{dixon1996sydney,harkema2009context,lee2015histology}.
A valid extraction therefore requires specimen attribution, assertion-status interpretation, and diagnostic-relation assessment rather than keyword detection alone.

These characteristics mean that existing clinical information-extraction systems cannot be assumed to transfer unchanged to a new pathology task.
Keyword and rule-based systems can encode terminology, report sections, and negation, but their rules generally require revision when local reporting conventions or requested variables change \cite{wang2018clinicalie,savova2010ctakes}.
Supervised extractors are tied to predefined labels and annotated training examples, so newly requested fields or institutional differences may require additional annotation and target-domain adaptation \cite{mitchell2022cancerbert,shimizu2025clinicaladaptation}.
Prompted LLMs reduce the need for task-specific training, but their outputs remain conditioned on field definitions, examples, prompting choices, and output schemas, and a plausible value alone does not guarantee correct specimen attribution, standardized report-level aggregation, or verifiable source evidence \cite{truhn2024gpt4pathology,balasubramanian2025llmpathology,wong2025universalabstraction}.
The remaining need is therefore not simply another pathology classifier, but a configurable workflow that can operationalize clinician-defined fields, accommodate different contextual requirements, and return consistent outputs with reviewable supporting evidence.

nMAS addresses these requirements through a configurable, evidence-linked workflow.
Each clinician-defined field is represented by a versioned field specification containing its intended meaning, output structure, clinical context, examples, and guardrails.
The Achievability Agent rejects requests that cannot be supported by the available report or configured schema, the Query Parser routes direct lexical and context-dependent fields to different extraction agents, FE-MUX assembles field-level predictions into a consistent report-level record, and the Output Validation Agent checks schema compliance and source-text grounding.
The distinctive contribution is therefore not the use of multiple agents alone or an assumption of predictive superiority, but the implementation of task adaptation, complexity-based routing, report-level aggregation, and evidence validation as explicit reusable stages within one clinician-reviewable workflow.

We retrospectively evaluated nMAS on 54 dummy gastric biopsy pathology reports from a large healthcare system in Singapore. Four clinician-scoped binary fields---gastric/stomach biopsy, biopsy status, \textit{H. pylori} positivity, and \textit{H. pylori}-associated gastritis---yielded 216 feature-case decisions. We also implemented a single-model UMA-style comparator to contextualize label-level performance. Because both approaches achieved similar classification performance, the study evaluates whether strong extraction can be maintained within a configurable and traceable workflow rather than claiming that a multi-agent architecture is intrinsically more accurate.

\section{Related Work}

Clinical information extraction has progressed from keyword and rule-based systems to supervised transformers and prompt-based large language models (LLMs).
Keyword and rule-based methods can encode dictionaries, report sections, assertion status, and negation, but they require task-specific maintenance as terminology, document structure, and target definitions change \cite{wang2018clinicalie,savova2010ctakes}.
Supervised models and domain-specific encoders such as BioBERT capture broader context, yet they depend on annotated target-domain data and may require adaptation across institutions with different formatting and vocabulary \cite{lee2020biobert,shimizu2025clinicaladaptation}.
LLM-based approaches reduce the need for task-specific training: UniMedAbstractor uses configurable prompts for clinical attribute extraction, and recent studies have evaluated zero-shot or prompted extraction from cancer pathology reports \cite{wong2025universalabstraction,truhn2024gpt4pathology,balasubramanian2025llmpathology}.

However, these approaches solve different parts of the extraction problem and remain coupled to their target definitions and evaluation settings.
Rule-based systems can encode local terminology, sections, and negation patterns, but their rules must be revised when reporting conventions or requested variables change.
Supervised systems can learn contextual patterns beyond exact matching, but they remain dependent on annotated examples for a predefined label set and may require target-domain adaptation across institutions.
Prompted LLMs reduce the need for retraining, but their predictions remain sensitive to field definitions, examples, output instructions, and the handling of unsupported or malformed responses \cite{wang2018clinicalie,savova2010ctakes,mitchell2022cancerbert,shimizu2025clinicaladaptation,truhn2024gpt4pathology,balasubramanian2025llmpathology,wong2025universalabstraction}.

Pathology-specific and gastrointestinal systems further demonstrate that extraction performance is generally evaluated within a particular organ, report type, and target schema.
Prior work has extracted breast cancer variables, Barrett's esophagus dysplasia, gastric disease categories, colonoscopy quality indicators, and predefined cancer-registry attributes \cite{achilonu2022rulebased,wenker2023barrettnlp,song2022gastricnlp,bae2022colonoscopy_nlp,aalabdulsalam2026multiagent}.
These systems can be effective within their intended scope, but their outputs do not automatically satisfy the requirements of a new gastric pathology request.
For example, a primary-care symptom extractor does not represent specimen-level report structure, an endoscopy pipeline does not necessarily distinguish organism detection from a pathologist's diagnostic attribution, and a cancer-registry extractor configured for fixed tumor variables does not automatically support newly defined \textit{H. pylori} fields.
Direct transfer therefore requires re-specifying the target semantics, adapting terminology and section logic, and implementing task-appropriate output and validation rules.

Most closely related, Aal Abdulsalam et al. proposed a modular multi-agent workflow for extracting predefined cancer-registry variables from pathology and medical reports \cite{aalabdulsalam2026multiagent}.
Their work establishes the feasibility of multi-agent pathology extraction for a defined cancer-registry schema.
nMAS addresses a complementary workflow problem by treating adaptation to a clinician-defined extraction request as part of the system itself: field specifications are configurable, unsupported requests are screened before extraction, fields are routed according to contextual complexity, field-level outputs are consolidated into a report-level structure, and retained predictions are validated against verbatim source evidence.

\section{Data and Methods}

\noindent\textbf{Dataset Overview.} This pilot evaluation used 54 dummy gastric biopsy pathology reports from 2022 from the anatomical pathology department of a large tertiary healthcare system in Singapore.
The sample was intended for an initial technical evaluation of the extraction workflow rather than to represent the institution's full pathology volume or estimate population prevalence.

Each report-level record included dummy administrative metadata, coded specimen and diagnosis fields, procedure and sign-out information, and the complete pathology report text.
Direct patient identifiers were removed or replaced with dummy placeholders before analysis, while clinically relevant wording required for extraction was preserved.

The reports varied in specimen labeling, diagnostic phrasing, organism descriptions, negation patterns, microscopic descriptions, and ancillary-test wording.
Relevant evidence could occur across both coded and narrative fields, requiring the extraction workflow to process the complete report record.
Table~\ref{tab:data_examples} presents two representative excerpts illustrating multi-specimen attribution and explicit negation, while a complete dummy report corresponding to Case~A is provided in Appendix~\ref{app:sample_report}.

\begin{table*}[t]
\centering
\caption{Representative dummy excerpts illustrating multi-specimen attribution and negated organism findings.}
\label{tab:data_examples}
\footnotesize
\setlength{\tabcolsep}{4pt}
\renewcommand{\arraystretch}{1.08}
\begin{tabularx}{\textwidth}{@{}>{\RaggedRight\arraybackslash}p{0.035\textwidth}>{\RaggedRight\arraybackslash}X@{}}
\toprule
\textbf{Case} & \multicolumn{1}{c}{\textbf{dummy Report Excerpt}} \\
\midrule
A & ``\textbf{Diagnosis:} (1) Rectosigmoid polyp: Tubular adenoma with low grade dysplasia. (2) Stomach; biopsy: Severe chronic acute antral and body gastritis with mild colonization by \textit{Helicobacter pylori}. There is no intestinal metaplasia, dysplasia or malignancy.'' \\
\midrule
B & ``\textbf{Diagnosis:} Gastric biopsy: Chronic gastritis. No Helicobacter organisms are identified. Negative for intestinal metaplasia, dysplasia and malignancy. \textbf{Microscopic description:} Helicobacter organisms are absent.'' \\
\bottomrule
\end{tabularx}
\end{table*}

\noindent\textbf{Target Features.} 
\label{sec:target_features}
Four binary target features were selected through consultation with four clinicians, including senior pathologists: gastric/stomach biopsy, biopsy status, \textit{H. pylori} positivity, and \textit{H. pylori}-associated gastritis. The first two fields define the target specimen cohort, whereas the latter two distinguish organism detection from an explicit diagnostic association between \textit{H. pylori} and gastritis.

\begin{enumerate}[leftmargin=*,label=\arabic*.,topsep=2pt,itemsep=1pt,parsep=0pt,partopsep=0pt]
\item \textbf{Gastric/Stomach Biopsy:} whether the report identifies a gastric or stomach biopsy specimen.
\item \textbf{Biopsy:} whether the specimen is a biopsy rather than another specimen type.
\item \textbf{\textit{H. pylori} Positive:} whether \textit{H. pylori}, Helicobacter organisms, or Helicobacter-like organisms are identified.
\item \textbf{\textit{H. pylori} Gastritis:} whether gastritis is explicitly associated with \textit{H. pylori}.
\end{enumerate}

\noindent\textbf{Reference Standard and Evidence Annotation.} \label{sec:reference_standard} 
Clinician-reviewed reference labels were defined for each report and each of the four target features using the complete dummy report record, including coded fields, specimen labels, final diagnosis text, microscopic descriptions, organism-related statements, and ancillary stain comments where available. A second reviewer independently reviewed all 216 feature-case labels, and inter-rater reliability between the two reviewers was quantified using Cohen's $\kappa$ before adjudication. Final reviewed labels were used as the reference standard. Reference evidence spans were recorded, where applicable, to document the source text supporting each reference label.

\begin{figure*}[!t]
\centering
\makebox[\textwidth][c]{%
\includegraphics[
    width=1.10\textwidth,
    trim=10pt 10pt 10pt 8pt,
    clip
]{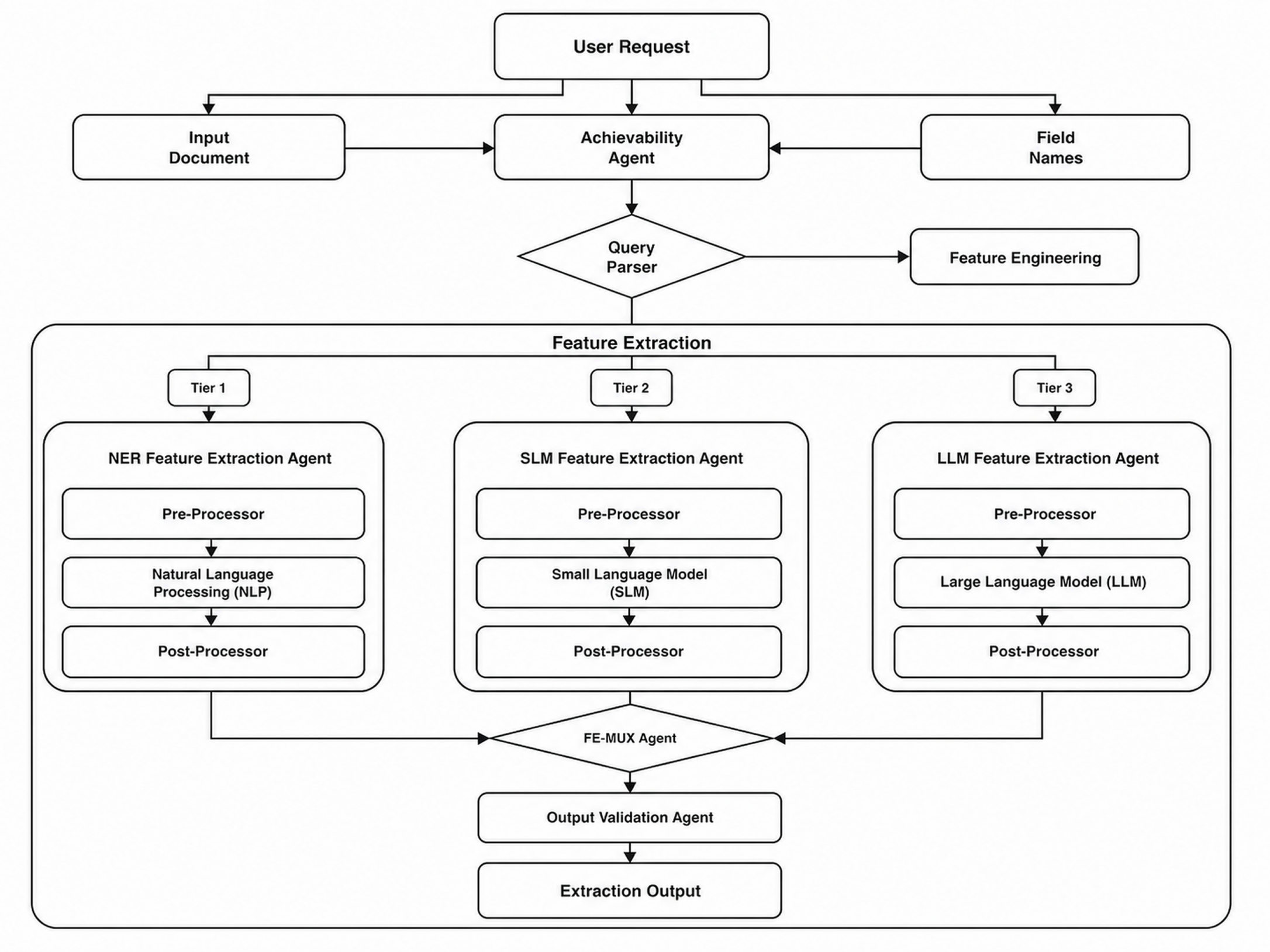}
}
\caption{Overview of the nMAS field-name driven extraction workflow. Target fields are routed through complexity-ranked extraction tiers, merged, validated against source-text evidence, and returned as structured outputs.}
\label{fig:pipeline}
\end{figure*}

\noindent\textbf{Workflow Overview.} As shown in Fig.~\ref{fig:pipeline}, nMAS supports both feature engineering and feature extraction, although only the extraction pathway was evaluated in this study. The workflow assesses whether each request is supported, routes fields according to contextual complexity, performs tier-specific extraction, consolidates outputs through FE-MUX, and validates supporting evidence against the source report.

\noindent\textbf{Input Standardization and Field Preparation.} Before extraction, each report record was converted into a consistent textual representation. Formatting differences involving line breaks, spacing, section separators, and repeated administrative headers were normalized where appropriate. Clinically relevant content was preserved, including coded specimen information, specimen labels, final diagnosis text, microscopic descriptions, organism-related statements, ancillary test results, and negation.

Each target field was paired with an entry in the configurable \texttt{FIELD\_LIBRARY}. Each entry specified the target meaning, expected output format, clinical context, in-context demonstrations, and field-specific guardrails. These entries operationalized the clinician-informed field definitions described in Section~\ref{sec:target_features} and guided field interpretation at inference time.

\noindent\textbf{Achievability Agent and Query Parsing.} The Achievability Agent assessed whether the input contained readable pathology content and whether each requested field was represented in the configured extraction schema. Requests lacking usable report text or a valid field definition were returned as unsupported rather than being assigned an inferred value.

For supported requests, the Query Parser acted as a routing agent. It constructed a field-specific extraction request containing the report text, target field, corresponding field definition, and expected output schema. It then assigned the request to one of the downstream extraction agents according to the configured linguistic and reasoning requirements of the field. Routing did not use the reference labels.

\noindent\textbf{Tiered Feature Extraction and Prompt Design.} After query parsing, each supported field request was assigned to one of three extraction routes according to its configured linguistic and reasoning requirements. Each route was implemented as a specialized extraction agent with its own pre-processing, extraction, and post-processing components.

Tier~1, the Named Entity Recognition Feature Extraction Agent, was used for fields supported primarily by direct lexical or coded evidence. This route consisted of a pre-processor, a natural language processing extraction component, and a post-processor that converted the result into the required format.

Tier~2, the Small Language Model Feature Extraction Agent, and Tier~3, the Large Language Model Feature Extraction Agent, were used for fields requiring contextual interpretation. Model assignments were determined before the present evaluation through engineering experiments using the same tiered nMAS workflow on a separate pathology information-extraction task. The broader candidate screening included the models ultimately selected for deployment, Qwen2.5-7B-Instruct and DeepSeek-V4-Flash, together with additional candidate models including GLM-5, Gemini~3.1~Pro, DeepSeek~V3.2, Kimi~K2~Thinking, and Qwen3-Next-80B \cite{yang2024qwen25,deepseekai2026deepseekv4,glm5_2026, google2026gemini31pro,deepseekai2025deepseekv32, moonshotai2025kimik2thinking,qwen2025qwen3next}. Models were compared based on field-level extraction performance, structured-output reliability, source-text grounding, negation handling, and runtime efficiency.

Following this broader screening, DeepSeek-V4-Flash \cite{deepseekai2026deepseekv4} was evaluated in the deployment workflow and achieved the best overall performance across these criteria. It was therefore selected for Tier~3, whereas Qwen2.5-7B-Instruct \cite{yang2024qwen25} was selected for Tier~2. Neither the 54 reports in the present study nor their reference labels were used for model selection or training.

Tier~2 was used for bounded contextual extraction, whereas Tier~3 was used for fields requiring broader contextual interpretation, including assertion-status assessment, negation handling, and the diagnostic association between \textit{H. pylori} and gastritis. Each language-model route used a pre-processor to combine the report text with the relevant field definition, a model component to extract the requested information, and a post-processor to normalize the result into the required schema.

The shared extraction instruction required report-only extraction, omission of absent or unclear values, structured JSON output, and verbatim supporting evidence.
Field-specific interpretation was supplied through the corresponding \texttt{FIELD\_LIBRARY} entry.
The complete core instruction and \textit{H. pylori}-specific field guidance are provided in Appendix~\ref{app:hpy_prompt}.

These global guardrails applied to all target fields.
They prohibited unsupported inference, required direct source-text evidence for every retained value, preserved conflicting supported mentions for review, and enforced the predefined JSON schema.
When evidence appeared in multiple sections, definitive diagnostic interpretation was prioritized over less authoritative content such as clinical history or administrative metadata.

\noindent\textbf{Field-Specific In-Context Demonstrations and Guardrails.} Each field-specific demonstration paired a short pathology-style excerpt with an expected field value and a guardrail defining the relevant extraction boundary. As illustrated in Table~\ref{tab:hpy_icl_representative}, the demonstrations addressed gastric and stomach terminology, common biopsy abbreviations, affirmative and negated organism findings, ancillary-stain evidence, and the requirement for an explicit diagnostic association before assigning \textit{H. pylori}-associated gastritis. The complete set of demonstrations is provided in Appendix Table~\ref{tab:hpy_icl_complete}.

\noindent\textbf{Feature Extraction Multiplexing.} Outputs from the active extraction agents were passed to the Feature Extraction Multiplexer Agent (FE-MUX), which assembled the field-level results into a single report-level structure. The merged representation retained each predicted value together with its source sentence, date, confidence score, and explanatory notes where available. Schema compliance and source-grounding checks were performed by the subsequent validation stage.

\begin{table*}[!t]
\centering
\caption{Representative \textit{H. pylori}-Specific ICL Demonstrations and Field-Mapping Guardrails}
\label{tab:hpy_icl_representative}
\scriptsize
\setlength{\tabcolsep}{3pt}
\renewcommand{\arraystretch}{1.15}
\begin{tabular}{
p{0.17\textwidth}
p{0.34\textwidth}
p{0.27\textwidth}
p{0.16\textwidth}}
\toprule
\textbf{Target Feature} &
\textbf{Pathology-Style Demonstration Excerpt} &
\textbf{Expected Field Mapping} &
\textbf{Guardrail Demonstrated}
\tabularnewline
\midrule

Gastric Biopsy &
``Stomach; biopsy.'' &
\texttt{\detokenize{gastric_biopsy = Y}} when the specimen is explicitly from stomach or gastric tissue. &
Accept stomach and gastric as equivalent site terms.
\tabularnewline
\midrule

Biopsy &
``GASTRIC BX.'' &
\texttt{\detokenize{biopsy = Y}} when biopsy is expressed using an accepted abbreviation such as ``BX.'' &
Accept common biopsy abbreviations.
\tabularnewline
\midrule

\textit{H. pylori} Positive &
``No Helicobacter organisms are identified.'' &
\texttt{\detokenize{h_pylori_positive = N}} when the organism statement is explicitly negated. &
Negation overrides keyword presence.
\tabularnewline
\midrule

\textit{H. pylori} Gastritis &
``Helicobacter pylori associated active chronic gastritis.'' &
\texttt{\detokenize{h_pylori_gastritis = Y}} when gastritis is explicitly associated with \textit{H. pylori}. &
Require explicit diagnostic association.
\tabularnewline

\bottomrule
\end{tabular}
\end{table*}

\noindent\textbf{Output Validation and Generation.} The Output Validation Agent assessed both output structure and source grounding. It checked that each result followed the required schema, that the reported source sentence appeared verbatim in the input report, and that the sentence supported the predicted value in context.

Positive labels required affirmative source evidence. Negative labels required explicit absence or negation rather than lack of mention. For \textit{H. pylori}-associated gastritis, evidence of organism positivity alone was insufficient without diagnostic evidence linking \textit{H. pylori} to gastritis.

Missing, unsupported, or internally inconsistent outputs were flagged rather than accepted as fully grounded predictions. The generation stage then assembled the validated fields into the final report-level output for clinician review.

\noindent\textbf{External UMA-Style Benchmark.}
To contextualize label-level performance, we implemented a UMA-style one-attribute benchmark based on prompting strategy of \cite{wong2025universalabstraction}. The benchmark used the same 54 reports and four target features as nMAS. For each report-feature pair, MiniMax M2.5 received the full report text, target name, concise definition, field-specific guidance, and positive and negative examples, and returned one binary value with verbatim source evidence in structured JSON.
Reference labels, nMAS outputs, and correctness indicators were excluded so that the benchmark measured independent extraction.
MiniMax M2.5 was selected because it produced reliable structured outputs in preliminary checks \cite{minimax2026m25}.

\noindent\textbf{Reference Standard and Evaluation Metrics.} Extraction performance was evaluated using the clinician-reviewed reference labels and evidence annotations described in the \textbf{Reference Standard and Evidence Annotation} paragraph above. Each of the 54 reports contributed one prediction for each of the four target fields, yielding 216 feature-case evaluations for nMAS and 216 feature-case evaluations for the UMA-style MiniMax benchmark.

A prediction was counted as correct when its binary value matched the corresponding reference label. Missing, invalid, unsupported, or non-parseable predictions were counted as incorrect rather than being converted into negative labels. Accuracy was calculated separately for each field and across all feature-case evaluations. Positive-class precision, recall, and F1 score were also calculated using the clinician-reviewed labels as the reference standard. As a descriptive grounding check, we also verified whether the source sentences associated with correctly classified predictions appeared verbatim in the corresponding input reports.

\noindent\textbf{Tier-3 Prompt-Component Ablation.} To examine the contribution of field-specific prompt guidance, we conducted a paired ablation experiment on the two Tier~3 fields: \textit{H. pylori} positivity and \textit{H. pylori}-associated gastritis. These fields were selected for ablation because they were determined by clinical experts to be the most clinically relevant context-dependent targets for \textit{H. pylori}-related case finding. Unlike the specimen-related fields, they require interpretation of assertion status, negation, ancillary-stain findings, and the explicit diagnostic association between \textit{H. pylori} and gastritis.

The full condition used the shared Tier~3 extraction prompt together with the complete field-specific \texttt{FIELD\_LIBRARY} entries, whereas the ablated condition omitted the field-specific definitions, aliases, clinical context, guardrails, and in-context demonstrations.

For this ablation analysis, reports containing exact textual overlap with the in-context demonstrations were excluded to prevent demonstration wording from affecting the comparison between conditions. Both conditions were therefore evaluated on the same remaining 41 reports, yielding 82 paired feature-case decisions per condition.

\suppressfloats[t]

\section{Results}

Cohen's $\kappa$ was 0.975, indicating high reviewer agreement.
nMAS correctly classified 213 of 216 feature-case decisions, corresponding to an overall accuracy of 98.61\% (Table~\ref{tab:results}).
Gastric/stomach biopsy identification and biopsy status each achieved 100.00\% accuracy, whereas \textit{H. pylori} positivity and \textit{H. pylori}-associated gastritis achieved 98.15\% and 96.30\%, respectively.
The three errors comprised one false positive for \textit{H. pylori}-associated gastritis and two false negatives from one report, one for each \textit{H. pylori}-related field.

\begin{table*}[!t]
\centering
\caption{nMAS and External UMA-Style MiniMax M2.5 Performance Across 54 Gastric Biopsy Pathology Reports}
\label{tab:results}
\scriptsize
\setlength{\tabcolsep}{4pt}
\renewcommand{\arraystretch}{1.12}
\resizebox{\textwidth}{!}{%
\begin{tabular}{llcccccc}
\toprule
\textbf{Method} & \textbf{Feature} & \textbf{N} & \textbf{Correct} & \textbf{Acc.} & \textbf{Prec.} & \textbf{Rec.} & \textbf{F1} \\
\midrule
nMAS & Gastric/stomach biopsy & 54 & 54 & 100.00\% & 100.00\% & 100.00\% & 100.00\% \\
nMAS & Biopsy & 54 & 54 & 100.00\% & 100.00\% & 100.00\% & 100.00\% \\
nMAS & \textit{H. pylori} positive & 54 & 53 & 98.15\% & 100.00\% & 96.43\% & 98.18\% \\
nMAS & \textit{H. pylori} gastritis & 54 & 52 & 96.30\% & 96.30\% & 96.30\% & 96.30\% \\
\midrule
UMA + MiniMax M2.5 & Gastric/stomach biopsy & 54 & 54 & 100.00\% & 100.00\% & 100.00\% & 100.00\% \\
UMA + MiniMax M2.5 & Biopsy & 54 & 54 & 100.00\% & 100.00\% & 100.00\% & 100.00\% \\
UMA + MiniMax M2.5 & \textit{H. pylori} positive & 54 & 53 & 98.15\% & 100.00\% & 96.43\% & 98.18\% \\
UMA + MiniMax M2.5 & \textit{H. pylori} gastritis & 54 & 52 & 96.30\% & 96.30\% & 96.30\% & 96.30\% \\
\midrule
nMAS overall & Feature-case level & 216 & 213 & 98.61\% & 99.38\% & 98.77\% & 99.08\% \\
UMA + MiniMax M2.5 overall & Feature-case level & 216 & 213 & 98.61\% & 99.38\% & 98.77\% & 99.08\% \\
\bottomrule
\end{tabular}%
}
\end{table*}

The external UMA-style MiniMax M2.5 comparator matched nMAS at the label level. Both methods correctly classified 213 of 216 feature-case decisions (98.61\%) and produced the same three errors in the two context-dependent H. pylori-related fields. Thus, this pilot evaluation did not demonstrate a label-accuracy advantage for nMAS over the single-model comparator. The comparator results instead provide context for assessing whether the additional workflow components preserve strong extraction performance while supporting source-grounded validation and standardized report-level delivery.

\noindent\textbf{Representative Evidence-Linked Outputs.}
In addition to binary feature labels, nMAS returned verbatim source evidence for field-level review.
Among the 213 correctly classified feature-case decisions, all associated source spans were confirmed to occur verbatim in the corresponding pathology reports.
Table~\ref{tab:representative_evidence} presents one representative evidence-linked output for each target feature.

\begin{table*}[t]
\centering
\caption{Representative evidence-linked outputs for the four target features.
Each source excerpt is reproduced verbatim from a dummy pathology report.}
\label{tab:representative_evidence}
\small
\setlength{\tabcolsep}{6pt}
\renewcommand{\arraystretch}{1.18}

\begin{tabularx}{\textwidth}{
    >{\RaggedRight\arraybackslash}p{0.22\textwidth}
    >{\centering\arraybackslash}p{0.07\textwidth}
    >{\RaggedRight\arraybackslash}X
}
\toprule
\textbf{Target Feature} &
\textbf{Value} &
\textbf{Verbatim Source Evidence} \\
\midrule

Gastric/stomach biopsy &
Y &
``Stomach; biopsy. Severe chronic acute antral and body gastritis with mild colonization by Helicobacter pylori.'' \\

\addlinespace[2pt]

Biopsy &
Y &
``Gastric biopsy. It consists of 3 pieces of tissue measuring 0.1 cm to 0.3 cm in greatest dimension.'' \\

\addlinespace[2pt]

\textit{H. pylori} positive &
N &
``No Helicobacter organisms are identified. Helicobacter organisms are absent.'' \\

\addlinespace[2pt]

\textit{H. pylori}-associated gastritis &
Y &
``Gastric biopsy: Helicobacter pylori-associated active chronic gastritis. Helicobacter organisms are identified.'' \\

\bottomrule
\end{tabularx}
\end{table*}

\noindent\textbf{Tier-3 Field-Library Ablation.}
After excluding reports with exact ICL overlap, both the full and ablated conditions correctly classified 81 of 82 paired decisions, corresponding to 98.78\% accuracy and 98.88\% positive-class F1.
Both conditions produced identical per-field outcomes and the same single error for \textit{H. pylori}-associated gastritis.
Thus, removing the field-specific definitions, guardrails, and demonstrations produced no measurable label-level change in this leakage-controlled pilot.

\section{Discussion}

The comparable label-level performance of nMAS and the UMA-style comparator may show that strong extraction performance can be maintained within a configurable, source-grounded workflow designed for clinician review.
The contribution of nMAS therefore lies in workflow integration and traceability rather than predictive superiority.

The observed error pattern mirrors the pathology-specific challenges described in the Introduction.
The two specimen-related fields were supported by explicit coded or lexical evidence and were extracted correctly in all 54 reports, whereas all observed errors involved the context-dependent distinction between organism detection and diagnostic attribution.
This distinction cannot be resolved reliably from keyword presence alone because the same organism terminology may occur in affirmative or negated statements, and an affirmative organism finding does not necessarily establish that the accompanying gastritis is \textit{H. pylori}-associated.
The results therefore illustrate why a system developed for symptoms, radiology findings, endoscopy indicators, or a fixed cancer-registry schema may perform well within its original task yet still require new field definitions, assertion rules, specimen-context handling, and evidence validation for gastric pathology.

nMAS addresses this adaptation requirement by making these elements explicit and modular.
The \texttt{FIELD\_LIBRARY} records the clinical semantics, aliases, examples, and guardrails for each requested field.
Complexity-based routing separates fields supported by direct lexical evidence from those requiring contextual interpretation, FE-MUX preserves a consistent report-level output, and the validation stage links each retained value to source text.
The comparable label-level performance of nMAS and the UMA-style comparator shows that these workflow components did not confer a predictive advantage in this small dataset.
Their intended value is instead to make a new pathology extraction request configurable, reviewable, and auditable without redesigning the entire clinician-facing workflow.
The unchanged Tier~3 ablation performance should likewise be interpreted in the context of the small and relatively constrained dataset, in which many reports contained explicit diagnostic wording or negation, rather than as evidence that field-specific guidance is unnecessary for broader pathology schemas.

Operational estimates suggest that evidence-linked extraction could reduce repeated manual abstraction.
For 1,000 reports, five minutes of manual review per report would require 83.3 staff-hours, compared with approximately 1.4 hours at five seconds of verification per report.
This difference corresponds to approximately 81.9 staff-hours or USD~6,100 in potential staff-time value under a rounded hourly cost of USD~75 \cite{jobstreet2026crc,moh2025healthcare_salaries,salaryexpert2026pathologist,wise2026sgdusd}.
These values are illustrative rather than prospectively measured.

\noindent\textbf{Limitations.}
This pilot study included only 54 reports from one institution and four binary fields.
It did not prospectively measure clinician review time, implementation effort, latency, or the semantic completeness of returned evidence spans.
The broader workflow implications should therefore not be interpreted as evidence of generalizability across institutions, specialties, or document types.
Future work should evaluate larger multi-institutional datasets, more complex schemas, evidence-span quality, adaptation effort, and clinician usability \cite{vasey2022decideai}.
Until such validation is completed, nMAS should remain an evidence-linked case-finding and review-support workflow rather than a replacement for pathologist review or clinical judgment.

\bibliography{pathology_fe_refs}

@article{wenker2023barrettnlp,
  title={Using Natural Language Processing to Automatically Identify Dysplasia in Pathology Reports for Patients With Barrett's Esophagus},
  author={Wenker, Theresa Nguyen and Natarajan, Yamini and Caskey, Kadon and Novoa, Francisco and Mansour, Nabil and Pham, Huy Anh and Hou, Jason K. and El-Serag, Hashem B. and Thrift, Aaron P.},
  journal={Clinical Gastroenterology and Hepatology},
  volume={21},
  number={5},
  pages={1198--1204},
  year={2023},
  doi={10.1016/j.cgh.2022.09.005}
}

@article{wang2018clinicalie,
  title={Clinical information extraction applications: A literature review},
  author={Wang, Yanshan and Wang, Liwei and Rastegar-Mojarad, Majid and Moon, Sungrim and Shen, Feichen and Afzal, Naveed and Liu, Sijia and Zeng, Yuqun and Mehrabi, Saeed and Sohn, Sunghwan and others},
  journal={Journal of Biomedical Informatics},
  volume={77},
  pages={34--49},
  year={2018},
  doi={10.1016/j.jbi.2017.11.011},
}

@article{malfertheiner2022maastricht,
  title={Management of {Helicobacter pylori} infection: the {Maastricht VI/Florence} consensus report},
  author={Malfertheiner, Peter and Megraud, Francis and Rokkas, Theodore and Gisbert, Javier P. and Liou, Jyh-Ming and Schulz, Christian and Gasbarrini, Antonio and Hunt, Richard H. and Leja, Marcis and O'Morain, Colm and others},
  journal={Gut},
  volume={71},
  number={9},
  pages={1724--1762},
  year={2022},
  doi={10.1136/gutjnl-2022-327745}
}

@article{chey2024acg,
  title={{ACG} Clinical Guideline: Treatment of {Helicobacter pylori} Infection},
  author={Chey, William D. and Howden, Colin W. and Moss, Steven F. and Morgan, Douglas R. and Greer, Katarina B. and Grover, Shilpa and Shah, Shailja C.},
  journal={American Journal of Gastroenterology},
  volume={119},
  number={9},
  pages={1730--1753},
  year={2024},
  doi={10.14309/ajg.0000000000002968},
}

@article{savova2010ctakes,
  title={Mayo clinical Text Analysis and Knowledge Extraction System ({cTAKES}): architecture, component evaluation and applications},
  author={Savova, Guergana K. and Masanz, James J. and Ogren, Philip V. and Zheng, Jiaping and Sohn, Sunghwan and Kipper-Schuler, Karin C. and Chute, Christopher G.},
  journal={Journal of the American Medical Informatics Association},
  volume={17},
  number={5},
  pages={507--513},
  year={2010},
  doi={10.1136/jamia.2009.001560},
}

@article{lee2020biobert,
  title={{BioBERT}: a pre-trained biomedical language representation model for biomedical text mining},
  author={Lee, Jinhyuk and Yoon, Wonjin and Kim, Sungdong and Kim, Donghyeon and Kim, Sunkyu and So, Chan Ho and Kang, Jaewoo},
  journal={Bioinformatics},
  volume={36},
  number={4},
  pages={1234--1240},
  year={2020},
  doi={10.1093/bioinformatics/btz682},
}

@article{truhn2024gpt4pathology,
  title={Extracting structured information from unstructured histopathology reports using generative pre-trained transformer 4 ({GPT-4})},
  author={Truhn, Daniel and Loeffler, Chiara M. L. and M\"uller-Franzes, Gustav and Nebelung, Sven and Hewitt, Katherine J. and Brandner, Sebastian and Bressem, Keno K. and Foersch, Sebastian and Kather, Jakob Nikolas},
  journal={The Journal of Pathology},
  volume={262},
  number={3},
  pages={310--319},
  year={2024},
  doi={10.1002/path.6232},
}

@article{balasubramanian2025llmpathology,
  title={Leveraging large language models for structured information extraction from pathology reports},
  author={Balasubramanian, Jeya Balaji and Adams, Daniel and Roxanis, Ioannis and Berrington de Gonzalez, Amy and Coulson, Penny and Almeida, Jonas S. and Garc\'ia-Closas, Montserrat},
  journal={Journal of Pathology Informatics},
  volume={19},
  pages={100521},
  year={2025},
  doi={10.1016/j.jpi.2025.100521}
}

@misc{wong2025universalabstraction,
  title={Universal Abstraction: Harnessing Frontier Models to Structure Real-World Data at Scale},
  author={Wong, Cliff and Preston, Sam and Liu, Qianchu and Gero, Zelalem and Bagga, Jass and Zhang, Sheng and Jain, Shrey and Zhao, Theodore and Gu, Yu and Xu, Yanbo and Kiblawi, Sid and Weerasinghe, Roshanthi and Leidner, Rom and Young, Kristina and Piening, Brian and Bifulco, Carlo and Naumann, Tristan and Wei, Mu and Poon, Hoifung},
  year={2025},
  eprint={2502.00943},
  archivePrefix={arXiv},
  primaryClass={cs.CL}
}

@article{wieneke2015validation,
title={Validation of natural language processing to extract breast cancer pathology procedures and results},
author={Wieneke, Arika E. and Bowles, Erin J. A. and Cronkite, David and Wernli, Karen J. and Gao, Hongyuan and Carrell, David and Buist, Diana S. M.},
journal={Journal of Pathology Informatics},
volume={6},
pages={38},
year={2015},
doi={10.4103/2153-3539.159215},
}

@article{lee2016eradication,
  title={Association Between {Helicobacter pylori} Eradication and Gastric Cancer Incidence: A Systematic Review and Meta-analysis},
  author={Lee, Yi-Chia and Chiang, Tsung-Hsien and Chou, Cheng-Kai and Tu, Yu-Kang and Liao, Wen-Chih and Wu, Ming-Shiang and Graham, David Y.},
  journal={Gastroenterology},
  volume={150},
  number={5},
  pages={1113--1124.e5},
  year={2016},
  doi={10.1053/j.gastro.2016.01.028},
}

@article{fock1997hpylori_singapore,
title={Helicobacter pylori infection: current status in Singapore},
author={Fock, K. M.},
journal={Annals of the Academy of Medicine, Singapore},
volume={26},
number={5},
pages={637--641},
year={1997}
}

@article{chew2017hpylori_singapore,
title={The diagnosis and management of H. pylori infection in Singapore},
author={Chew, Claire A. Z. and Lye, Tong Fong and Ang, Daphne and Ang, Tiing Leong},
journal={Singapore Medical Journal},
volume={58},
number={5},
pages={234--240},
year={2017},
doi={10.11622/smedj.2017037}
}

@article{ang2021hpylori_strategy_singapore,
title={Helicobacter pylori Treatment Strategies in Singapore},
author={Ang, Tiing Leong and Ang, Daphne},
journal={Gut and Liver},
volume={15},
number={1},
pages={13--18},
year={2021},
doi={10.5009/gnl19308}
}

@article{achilonu2022rulebased,
title={Rule-Based Information Extraction from Free-Text Pathology Reports Reveals Trends in South African Female Breast Cancer Molecular Subtypes and Ki67 Expression},
author={Achilonu, Okechinyere J. and Singh, Elvira and Nimako, Gideon and Eijkemans, Ren{'e} M. J. C. and Musenge, Eustasius},
journal={BioMed Research International},
volume={2022},
pages={6157861},
year={2022},
doi={10.1155/2022/6157861}
}

@article{song2022gastricnlp,
title={Natural Language Processing for Information Extraction of Gastric Diseases and Its Application in Large-Scale Clinical Research},
author={Song, Gyuseon and Chung, Su Jin and Seo, Ji Yeon and Yang, Sun Young and Jin, Eun Hyo and Chung, Goh Eun and Shim, Sung Ryul and Sa, Soonok and Hong, Moongi Simon and Kim, Kang Hyun and Jang, Eun and Lee, Chae Won and Bae, Jung Ho and Han, Hyun Wook},
journal={Journal of Clinical Medicine},
volume={11},
number={11},
pages={2967},
year={2022},
doi={10.3390/jcm11112967}
}

@misc{jobstreet2026crc,
  title = {Clinical Research Coordinator Salary in Singapore},
  author = {{JobStreet}},
  year = {2026},
  note = {Accessed: 2026-06-05. Reported average monthly salary range: SGD 3,700--4,200},
  url = {https://sg.jobstreet.com/career-advice/role/clinical-research-coordinator/salary}
}

@misc{moh2025healthcare_salaries,
  title = {Average and Median Salaries Earned by Public Sector Healthcare Workers from 2021 to 2025},
  author = {{Ministry of Health Singapore}},
  year = {2025},
  note = {Accessed: 2026-06-05. Reported median monthly base salary of public-sector doctors in 2024: approximately SGD 14,400},
}

@misc{salaryexpert2026pathologist,
  title = {Pathologist Salary in Singapore},
  author = {{ERI SalaryExpert}},
  year = {2026},
  note = {Accessed: 2026-06-05. Reported average hourly rate: SGD 143.64},

}

@misc{wise2026sgdusd,
  title = {Singapore Dollar to US Dollars Exchange Rate History},
  author = {{Wise}},
  year = {2026},
  note = {Accessed: 2026-06-05. Mid-market exchange rate on June 5, 2026: 1 SGD = USD 0.7745},
}

@article{ang2022clinical_audit_hpylori_sg,
title={Clinical audit of current {Helicobacter pylori} treatment outcomes in Singapore},
author={Ang, Tiing Leong and Lim, Kim Wei and Ang, Daphne and Wong, Yu Jun and Tan, Malcolm and Wong, Andrew Siang Yih},
journal={Singapore Medical Journal},
volume={63},
number={9},
pages={503--508},
year={2022},
doi={10.11622/smedj.2021105}
}

@article{bae2022colonoscopy_nlp,
title={Natural Language Processing for Assessing Quality Indicators in Free-Text Colonoscopy and Pathology Reports: Development and Usability Study},
author={Bae, Jung Ho and Han, Hyun Wook and Yang, Sun Young and Song, Gyuseon and Sa, Soonok and Chung, Goh Eun and Seo, Ji Yeon and Jin, Eun Hyo and Kim, Heecheon and An, DongUk},
journal={JMIR Medical Informatics},
volume={10},
number={4},
pages={e35257},
year={2022},
doi={10.2196/35257}
}

@article{bian2026learningutility,
  author  = {Bian, Jiang and Afshar, Majid and Scifres, Christina M. and Webber, Emily and Burton, David and Vawdrey, David and Wang, Fei and Melton, Genevieve B. and Shah, Nigam and Patzer, Rachel E. and Zhang, Yiye},
  title   = {The bottleneck was never data or algorithms: building a learning utility for AI-enabled learning health systems},
  journal = {npj Health Systems},
  volume  = {3},
  number  = {43},
  year    = {2026},
  doi     = {10.1038/s44401-026-00107-x},
  url     = {https://www.nature.com/articles/s44401-026-00107-x}
}

@article{vasey2022decideai,
  author  = {Vasey, Baptiste and Nagendran, Myura and Campbell, Bruce
             and Clifton, David A. and Collins, Gary S. and Denaxas, Spiros
             and Denniston, Alastair K. and Faes, Livia and Geerts, Bart
             and Ibrahim, Mudathir and Liu, Xiaoxuan and Mateen, Bilal A.
             and Mathur, Piyush and McCradden, Melissa D. and Morgan, Lauren
             and Ordish, Johan and Rogers, Campbell and Saria, Suchi
             and Ting, Daniel S. W. and Watkinson, Peter and Weber, Wim
             and Wheatstone, Peter and McCulloch, Peter
             and {{DECIDE-AI expert group}}},
  title   = {Reporting Guideline for the Early-Stage Clinical Evaluation
             of Decision Support Systems Driven by Artificial Intelligence:
             {DECIDE-AI}},
  journal = {Nature Medicine},
  volume  = {28},
  number  = {5},
  pages   = {924--933},
  year    = {2022},
  doi     = {10.1038/s41591-022-01772-9}
}

@article{yang2024qwen25,
  title   = {{Qwen2.5 Technical Report}},
  author  = {Yang, An and Yang, Baosong and Zhang, Beichen and
             Hui, Binyuan and Zheng, Bo and Yu, Bowen and
             Li, Chengyuan and Liu, Dayiheng and Huang, Fei and
             Wei, Haoran and others},
  journal = {arXiv preprint arXiv:2412.15115},
  year    = {2024},
  doi     = {10.48550/arXiv.2412.15115}
}

@article{deepseekai2026deepseekv4,
  title   = {{DeepSeek-V4}: Towards Highly Efficient Million-Token
             Context Intelligence},
  author  = {{DeepSeek-AI}},
  journal = {arXiv preprint arXiv:2606.19348},
  year    = {2026},
  doi     = {10.48550/arXiv.2606.19348}
}

@misc{minimax2026m25,
  author       = {{MiniMax}},
  title        = {{MiniMax M2.5: Built for Real-World Productivity}},
  year         = {2026},
  month        = feb,
  url          = {https://www.minimax.io/news/minimax-m25},
  note         = {Accessed: June 28, 2026}
}

@article{glm5_2026,
  author        = {{GLM-5 Team}},
  title         = {{GLM-5}: From Vibe Coding to Agentic Engineering},
  journal       = {arXiv preprint arXiv:2602.15763},
  year          = {2026},
  eprint        = {2602.15763},
  archivePrefix = {arXiv},
  primaryClass  = {cs.CL}
}

@misc{google2026gemini31pro,
  author       = {{Google}},
  title        = {{Gemini 3.1 Pro Preview}},
  year         = {2026},
  howpublished = {Google AI for Developers},
  note         = {Accessed: Jun. 30, 2026},
  url          = {https://ai.google.dev/gemini-api/docs/models/gemini-3.1-pro-preview}
}

@article{deepseekai2025deepseekv32,
  author        = {{DeepSeek-AI}},
  title         = {{DeepSeek-V3.2}: Pushing the Frontier of Open Large Language Models},
  journal       = {arXiv preprint arXiv:2512.02556},
  year          = {2025},
  eprint        = {2512.02556},
  archivePrefix = {arXiv},
  primaryClass  = {cs.CL}
}

@misc{qwen2025qwen3next,
  author       = {{Qwen Team}},
  title        = {{Qwen3-Next-80B-A3B}},
  year         = {2025},
  howpublished = {Official Qwen model release},
  note         = {Accessed: Jun. 30, 2026},
  url          = {https://qwen.ai/blog?from=research.latest-advancements-list&id=4074cca80393150c248e508aa62983f9cb7d27cd}
}

@misc{moonshotai2025kimik2thinking,
  author       = {{Moonshot AI}},
  title        = {Introducing {Kimi K2 Thinking}},
  year         = {2025},
  howpublished = {Official model documentation},
  note         = {Accessed: Jun. 30, 2026},
  url          = {https://moonshotai.github.io/Kimi-K2/thinking.html}
}

@article{dixon1996sydney,
  title={Classification and Grading of Gastritis: The Updated Sydney System},
  author={Dixon, Michael F. and Genta, Robert M. and Yardley, John H. and Correa, Pelayo},
  journal={The American Journal of Surgical Pathology},
  volume={20},
  number={10},
  pages={1161--1181},
  year={1996},
  doi={10.1097/00000478-199610000-00001}
}

@article{lee2015histology,
  title={Diagnosis of \textit{Helicobacter pylori} by Invasive Test: Histology},
  author={Lee, Ji Young and Kim, Nayoung},
  journal={Annals of Translational Medicine},
  volume={3},
  number={1},
  pages={10},
  year={2015},
  doi={10.3978/j.issn.2305-5839.2014.11.03}
}

@article{harkema2009context,
  title={ConText: An Algorithm for Determining Negation, Experiencer, and Temporal Status from Clinical Reports},
  author={Harkema, Henk and Dowling, John N. and Thornblade, Tyler and Chapman, Wendy W.},
  journal={Journal of Biomedical Informatics},
  volume={42},
  number={5},
  pages={839--851},
  year={2009},
  doi={10.1016/j.jbi.2009.05.002}
}

@inproceedings{aalabdulsalam2026multiagent,
  title={A Multi-Agent Open-Source {LLM} for Structured Cancer Registry Information Extraction from Pathology and Medical Reports},
  author={Aal Abdulsalam, Abdulrahman and Al Zaabi, Adhari and Jeeballah, Riham and El Keraby, Habiba},
  booktitle={BioNLP 2026},
  pages={531--551},
  year={2026},
  publisher={Association for Computational Linguistics},
  address={San Diego, California},
  doi={10.18653/v1/2026.bionlp-1.43}
}

@inproceedings{shimizu2025clinicaladaptation,
  title={Exploring {LLM} Annotation for Adaptation of Clinical Information Extraction Models under Data-sharing Restrictions},
  author={Shimizu, Seiji and Shohei, Hisada and Uno, Yutaka and Yada, Shuntaro and Wakamiya, Shoko and Aramaki, Eiji},
  booktitle={Findings of the Association for Computational Linguistics: ACL 2025},
  pages={14678--14694},
  year={2025},
  publisher={Association for Computational Linguistics},
  address={Vienna, Austria},
  doi={10.18653/v1/2025.findings-acl.757}
}

@article{buckley2012feasibility,
  title={The Feasibility of Using Natural Language Processing to Extract Clinical Information from Breast Pathology Reports},
  author={Buckley, Julliette M. and Coopey, Suzanne B. and Sharko, John and Polubriaginof, Fernanda and Drohan, Brian and Belli, Ahmet K. and Kim, Elizabeth M. H. and Garber, Judy E. and Smith, Barbara L. and Gadd, Michele A. and Specht, Michelle C. and Roche, Constance A. and Gudewicz, Thomas M. and Hughes, Kevin S.},
  journal={Journal of Pathology Informatics},
  volume={3},
  pages={23},
  year={2012},
  doi={10.4103/2153-3539.97788}
}

@article{mitchell2022cancerbert,
  title={A Question-and-Answer System to Extract Data From Free-Text Oncological Pathology Reports ({CancerBERT} Network): Development Study},
  author={Mitchell, Joseph Ross and Szepietowski, Phillip and Howard, Rachel and Reisman, Phillip and Jones, Jennie D. and Lewis, Patricia and Fridley, Brooke L. and Rollison, Dana E.},
  journal={Journal of Medical Internet Research},
  volume={24},
  number={3},
  pages={e27210},
  year={2022},
  doi={10.2196/27210}
}

\appendix

\section{Representative dummy Gastric Biopsy Report}
\label{app:sample_report}

The following dummy report illustrates the source-record structure used for extraction, including coded specimen fields, diagnostic text, and gross-description content.

\begin{lstlisting}[style=reportstyle]
Specimen ID: 22:AB00001
Receive Date: 2022-01-11
Patient Name: Patient 1
ID: 0000001
Race: A
Sex: B

TCode:
T57010 (Gastric biopsy)
T57010 (Gastric biopsy)
T59604 (Rectal biopsy)

MCode:
GA003 (Severe)
M43000|T57010 (CHRONIC GASTRITIS)
M82110 (Tubular adenoma, NOS)

Diagnosis (full report):
CPOE
CPOE MESSAGE RECEIVED: 11/01/22 1214
HISTOPATHOLOGY: Y
VETTED & ORDER FORM COMPLETED (DR ONLY): Y
SPECIMEN LABEL COMPLETED: Y

ORDERSET:
Routine Specimen:
1. Rectal sigmoid polyp
2. Gastric BX

CLINICAL DIAGNOSIS: anaemia
TIME: 11:56

DIAGNOSIS

(1) Rectosigmoid polyp
TUBULAR ADENOMA WITH LOW GRADE DYSPLASIA.

(2) Stomach; biopsy
SEVERE CHRONIC ACUTE ANTRAL AND BODY GASTRITIS
WITH MILD COLONIZATION BY HELICOBACTER PYLORI.

THERE IS NO INTESTINAL METAPLASIA, DYSPLASIA
OR MALIGNANCY.

GROSS DESCRIPTION

The specimens are received in formalin, labelled
with patient's data and designated as follows.

(A) Rectal sigmoid polyp
It consists of a piece of tissue measuring 0.4 cm
in greatest dimension.
(A1-inked blue; no reserve)

(B) Gastric biopsy
It consists of 3 pieces of tissue measuring
0.1 cm to 0.3 cm in greatest dimension.
(B1-inked yellow; no reserve)

The specimens were fixed in formalin for 6--72 hours.

Order Location: S42A

Procedure:
HT.SPECIALBX2
HP.HE EMBED

Signout: AP-ABC
\end{lstlisting}

\section{Core Extraction Prompt and \textit{H. pylori}-Specific Field Guidance}
\label{app:hpy_prompt}

\begin{prompt}[Core extraction instruction]
\label{prompt:assessment}
Use only information explicitly present in the report.
Do not infer missing facts.
Interpret the target feature using the field library, including its context, examples, and guardrails.
If the requested value is absent or unclear, omit the feature.
Every retained extraction must include a verbatim supporting source sentence, a confidence score, and explanatory notes.
\end{prompt}

\begin{prompt}[\textit{H. pylori}-specific field guidance]
\label{prompt:hpy_guidance}

\textbf{\texttt{h\_pylori\_positive}}

\textbf{Y:} The report affirmatively identifies \textit{H. pylori}, Helicobacter organisms, or Helicobacter-like organisms in the gastric specimen.

\textbf{N:} The report explicitly states absence or non-identification, including a negative \textit{H. pylori} immunostain.

\textbf{Rule:} Negation overrides keyword presence.

\textbf{\texttt{h\_pylori\_gastritis}}

\textbf{Y:} The report explicitly links gastritis to \textit{H. pylori} or Helicobacter organisms.

\textbf{N:} Gastritis is present, but \textit{H. pylori} is absent or is not diagnostically associated with the gastritis.

\textbf{Rule:} Gastritis alone and organism positivity alone are insufficient without an explicit diagnostic association.
\end{prompt}

\section{Complete H. pylori Specific ICL Demonstrations}
\label{app:full_icl}

Table~\ref{tab:hpy_icl_complete} shows the complete set of field-specific ICL demonstrations used to define the four target features. The demonstrations include multiple examples per feature and cover positive wording, negative wording, abbreviations, protocol-based biopsy descriptions, organism identification, ancillary stain evidence, and diagnostic association with gastritis.

\begin{table*}[!t]
\centering
\caption{Complete \textit{H. pylori}-Specific ICL Demonstrations and
Field-Mapping Guardrails}
\label{tab:hpy_icl_complete}
\scriptsize
\setlength{\tabcolsep}{3pt}
\renewcommand{\arraystretch}{1.12}

\begin{tabular}{
p{0.17\textwidth}
p{0.34\textwidth}
p{0.27\textwidth}
p{0.16\textwidth}}
\toprule
\textbf{Target Feature} &
\textbf{Pathology-Style Demonstration Excerpt} &
\textbf{Expected Field Mapping} &
\textbf{Guardrail Demonstrated}
\tabularnewline
\midrule

Gastric Biopsy &
``Stomach; biopsy.'' &
\texttt{\detokenize{gastric_biopsy = Y}} when the specimen is explicitly from stomach or gastric tissue. &
Accept stomach and gastric as equivalent site terms.
\tabularnewline
\midrule

Gastric Biopsy &
``GASTRIC ANTRUM BIOPSY.'' &
\texttt{\detokenize{gastric_biopsy = Y}} when the biopsy site is antrum, body, cardia, fundus, or another gastric subsite. &
Gastric subsite wording supports gastric biopsy.
\tabularnewline
\midrule

Gastric Biopsy &
``Stomach, Sydney protocol biopsy.'' &
\texttt{\detokenize{gastric_biopsy = Y}} when Sydney protocol biopsy is described as a stomach/gastric biopsy. &
Recognize protocol-based gastric sampling.
\tabularnewline
\midrule

Biopsy &
``Gastric biopsy. It consists of 3 pieces of tissue measuring 0.1 cm to 0.3 cm.'' &
\texttt{\detokenize{biopsy = Y}} when the specimen is explicitly described as biopsy tissue. &
Use specimen type and gross description.
\tabularnewline
\midrule

Biopsy &
''GASTRIC BX.'' &
\texttt{\detokenize{biopsy = Y}} when biopsy is expressed using an accepted abbreviation such as `BX.'' &
Accept common biopsy abbreviations.
\tabularnewline
\midrule

Biopsy &
``Random biopsies taken (Updated Sydney Protocol) antrum, incisura, and body.'' &
\texttt{\detokenize{biopsy = Y}} when the procedure states that biopsy samples were taken. &
Procedure text can support biopsy status.
\tabularnewline
\midrule

\textit{H. pylori} Positive &
``MILD COLONIZATION BY HELICOBACTER PYLORI.'' &
\texttt{\detokenize{h_pylori_positive = Y}} when the report affirmatively states colonization by \textit{H. pylori}. &
Affirmative organism evidence supports positivity.
\tabularnewline
\midrule

\textit{H. pylori} Positive &
``Helicobacter organisms are identified.'' &
\texttt{\detokenize{h_pylori_positive = Y}} when Helicobacter organisms are explicitly identified. &
Direct organism identification supports positivity.
\tabularnewline
\midrule

\textit{H. pylori} Positive &
``Helicobacter pylori-like organisms are highlighted by immunohistochemistry.'' &
\texttt{\detokenize{h_pylori_positive = Y}} when ancillary staining highlights \textit{H. pylori}-like organisms. &
Positive IHC evidence supports positivity.
\tabularnewline
\midrule

\textit{H. pylori} Positive &
``No Helicobacter organisms are identified.'' &
\texttt{\detokenize{h_pylori_positive = N}} when the organism statement is explicitly negated. &
Negation overrides keyword presence.
\tabularnewline
\midrule

\textit{H. pylori} Positive &
``No definite Helicobacter pylori identified.'' &
\texttt{\detokenize{h_pylori_positive = N}} when the report states that definite organisms are not identified. &
Do not infer positivity from uncertain or negative wording.
\tabularnewline
\midrule

\textit{H. pylori} Positive &
``No conspicuous Helicobacter pylori organisms are identified, corroborated by negative \textit{H. pylori} immunostain.'' &
\texttt{\detokenize{h_pylori_positive = N}} when negative morphology is supported by negative immunostain. &
Negative IHC supports a negative value.
\tabularnewline
\midrule

\textit{H. pylori} Gastritis &
``Helicobacter pylori associated active chronic gastritis.'' &
\texttt{\detokenize{h_pylori_gastritis = Y}} when gastritis is explicitly associated with \textit{H. pylori}. &
Require explicit diagnostic association.
\tabularnewline
\midrule

\textit{H. pylori} Gastritis &
``Moderate to marked \textit{H. pylori} associated active chronic gastritis.'' &
\texttt{\detokenize{h_pylori_gastritis = Y}} when severity is described together with \textit{H. pylori}-associated gastritis. &
Preserve the association even with severity modifiers.
\tabularnewline
\midrule

\textit{H. pylori} Gastritis &
``Helicobacter pylori-associated moderate chronic gastritis with focal activity.'' &
\texttt{\detokenize{h_pylori_gastritis = Y}} when the diagnosis links \textit{H. pylori} to chronic gastritis. &
Final diagnosis can support the association.
\tabularnewline
\midrule

\textit{H. pylori} Gastritis &
``Mild chronic gastritis. No Helicobacter organisms are identified.'' &
\texttt{\detokenize{h_pylori_gastritis = N}} when gastritis is present but organisms are explicitly absent. &
Gastritis alone is insufficient.
\tabularnewline
\midrule

\textit{H. pylori} Gastritis &
``Mild-to-moderate chronic gastritis. No definite Helicobacter pylori identified.'' &
\texttt{\detokenize{h_pylori_gastritis = N}} when chronic gastritis is not attributed to \textit{H. pylori}. &
Do not label non-associated gastritis as \textit{H. pylori} gastritis.
\tabularnewline
\midrule

\textit{H. pylori} Gastritis &
``Helicobacter organisms are absent.'' &
\texttt{\detokenize{h_pylori_gastritis = N}} when organism absence prevents attribution of gastritis to \textit{H. pylori}. &
Organism absence rules against \textit{H. pylori}-associated gastritis.
\tabularnewline

\bottomrule
\end{tabular}
\end{table*}

\section{Tier-3 Ablation Prompt Templates}
\label{app:tier3_prompts}

\begin{prompt}[Ablated Tier-3 baseline prompt]
\label{prompt:tier3_ablated}
You are a clinical NLP assistant specialised in parsing Singapore hospital
gastrointestinal pathology reports.

Extract the following two \textit{H. pylori} features from the report:
\texttt{h\_pylori\_pos} and \texttt{h\_pylori\_gastritis}. Use only the
pathology report text. Answer each feature with \texttt{Y} or \texttt{N}.
Return only valid JSON with the keys:
\texttt{h\_pylori\_pos}, \texttt{verbatim\_h\_pylori\_pos},
\texttt{h\_pylori\_gastritis},
\texttt{verbatim\_h\_pylori\_gastritis}, and \texttt{reasoning}.
\end{prompt}

\begin{prompt}[Full Tier-3 prompt with field-specific guidance]
\label{prompt:tier3_full}
Use the same baseline Tier-3 prompt shown in
Prompt~\ref{prompt:tier3_ablated}. In addition, apply the
field-specific \texttt{FIELD\_LIBRARY} entries for
\texttt{h\_pylori\_positive} and \texttt{h\_pylori\_gastritis}, including
their clinical context, guardrails, and in-context demonstrations. The
output key \texttt{h\_pylori\_pos} corresponds to the field-library entry
\texttt{h\_pylori\_positive}.
\end{prompt}

\end{document}